\documentclass{SIW}

\usepackage[OT1]{fontenc}
\usepackage{graphicx,xcolor}
\usepackage{lipsum}
\usepackage{amsmath}
\usepackage{amsfonts}
\usepackage{amssymb}
\usepackage{hyperref}
\usepackage{framed}
\usepackage{multirow}
\usepackage{booktabs}

\usepackage{ifthen}
\newboolean{draftenabled}
\setboolean{draftenabled}{true} 

\ifthenelse{\boolean{draftenabled}}
{
    \newenvironment{draft}{
       
       \MakeFramed{\advance\hsize-\width\FrameRestore}}
     {\endMakeFramed}

    \newcommand{\draftinline}[1]{\colorbox{pink}{#1}}
}{
    \newcommand{\draftinline}[1]{}

}

\newcommand{\eqtref}[1]{Eq.~\eqref{#1}}

\newcommand{\Eradar}{E_\text{radar}}
\newcommand{\Elores}{E_\text{lores}}
\newcommand{\Ehires}{E_\text{hires}}

\author{
	Patrick Kage\textsuperscript{1}\textsuperscript{*}
	and 
	Pavlos Andreadis\textsuperscript{1};
    \textsuperscript{1}Artificial Intelligence and its Applications Institute, (University of Edinburgh, 10 Crichton Street, Newington, Edinburgh EH8 9AB)
	\textsuperscript{*}[p.kage@ed.ac.uk]
}

\title{Multi-modal, multi-scale representation learning for satellite imagery analysis just needs a good ALiBi}

\event{Space Imaging Workshop. Atlanta, GA. \\ 7-9 October 2024}

\begin{document}

\maketitle

\MakeAbstract{
    Vision foundation models have been shown to be effective at processing
    satellite imagery into representations fit for downstream tasks, however,
    creating models which operate over multiple spatial resolutions and modes
    is challenging. This paper presents Scale-ALiBi, a linear bias transformer
    attention mechanism with a spatial encoding bias to relationships between
    image patches at different ground sample distance scales. We provide an
    implementation of Scale-ALiBi over a dataset of aligned high- and
    low-resolution optical and low-resolution SAR satellite imagery data using
    a triple-contrastive and reconstructive architecture, show an improvement
    on the GEO-Bench benchmark, and release the newly curated dataset publicly.
}

\section*{Introduction}

The volume of satellite imagery generated by both governmental and commercial
constellations has been increasing year-over-year, far eclipsing the ability
for human analysts to keep up. Satellite imagery presents an ideal
use-case for representation learning: while there is very little labeled data,
the images captured are pre-orthorectified and tagged with location
information, the sensors used to generate the images are well-characterized,
and individual scenes are frequently revisited. This means that for any given
location on Earth, there are multiple image captures both across sensor
modalities and throughout time which are easily machine-alignable.
Representation learning allows for the automatic extraction of meaningful
features from multi-modal satellite imagery and thus makes downstream tasks
such as land use classification and change monitoring simpler to implement.
Currently, representation learning models focus on learning stable
representations across ground sample distances (GSDs), or across image
modalities (e.g.~radar to equivalent-resolution optical), or across temporal
captures. However, few models attempt to capture more than one of these
representations at a time.

This paper's key contributions include first the extension of the
2D-ALiBi/X-ALiBi attention mechanism\cite{CROMA} with GSD scaling to allow
representation learning transformers to incorporate both scale and distance
information from satellite images into the training process, and second the
evaluation of the resulting attention mechanism over multi-scale multi-modal
imagery using a novel triple-contrastive architecture. This allows for the
creation of a representation model which operates natively over both
multi-modal and multi-resolution imagery. In order to train this
model, a new dataset of aligned image pairs is curated by the authors, sourcing
data from ESA's Sentinel-1 SAR (Synthetic-Aperture Radar) and Sentinel-2 MSI
(MultiSpectral Instrument) imaging missions\cite{Sentinel} and the U.S.
Department of Agriculture's National Agriculture Imagery Program (NAIP) high
resolution image acquisitions\cite{NAIP}. This dataset is released with the
paper to facilitate further research.

\section*{Background}

Foundation models in the satellite imagery representation learning space are
largely implemented as self-supervised vision
transformers\cite{satmae,ScaleMAE,CROMA}. Transformers, while originally
designed for natural language processing tasks, have been shown to be effective at
processing images once the input images are split into a sequence of patches
which are then processed similarly to language tokens\cite{vit}. When trained
in a self-supervised manner, these vision transformers learn representations
without requiring labeled information (which is expensive to acquire at scale).

Scale-MAE\cite{ScaleMAE} has demonstrated the effectiveness of scaling the
sinusoidal position encoding of image tokens by the GSD of the input sample,
explicitly learning the relationship between low- and high-resolution views of
a single modality sample. Similarly, CROMA\cite{CROMA} demonstrated that
contrastive learning can learn a cross-modal representation between Sentinel-1
synthetic-aperture radar (SAR) and Sentinel-2 MSI (optical) patches of uniform
size. CROMA also introduced an extension of the ALiBi linear bias attention
mechanism\cite{alibi} into two dimensions for both self-attention (2D-ALiBi)
and cross-attention (X-ALiBi), encoding the Euclidean distance between sample
pairs. Linear bias attention allows for the transformer to extrapolate to
sequences longer than sequences presented during training\cite{alibi}, which is
a desirable property in remote sensing, as images can be extremely large.

\section*{Method}

\begin{figure*}[ht!]
	\centering
	\includegraphics[width=0.90\textwidth]{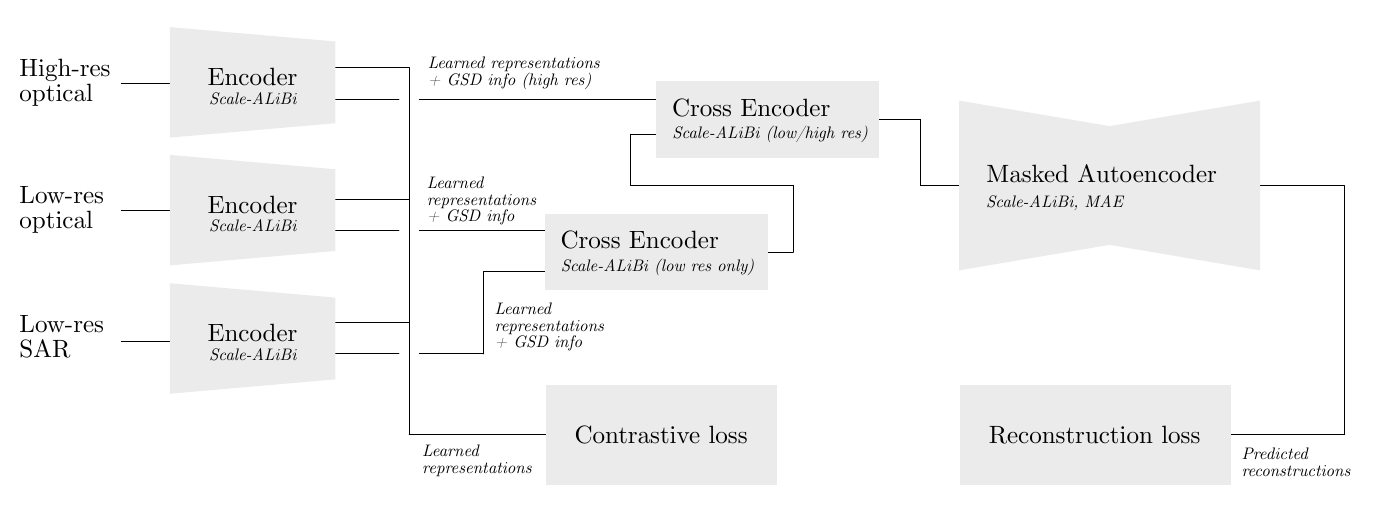}
    \vspace{0.2cm}
    \caption{A block diagram of the full Scale-ALiBi model.}
	\label{fig:arch}
\end{figure*}

\subsection*{Scale-ALiBi attention}

In order to add a GSD scale-aware component to our representation model, we
introduce the \textit{Scale-ALiBi} attention mechanism. We define the
Scale-ALiBi matrix very similarly to the 2D-ALiBi matrix, with an attention
matrix \(A \in \mathbb R^{h\times L\times L}\) for \(h\) heads with sequence
length \(L\) and head depth \(d\). Each position in the attention matrix is
given by \eqtref{eq:scalealibi_1}:

\begin{equation}\label{eq:scalealibi_1}
    a_{hij} = \underbrace{\sqrt{d} \cdot q_{hi} \cdot k_{hj} }_\text{normal
    attention} - \underbrace{g(i,j) \cdot m(h)}_\text{Scale-ALiBi}
\end{equation}

\noindent
where \(q_{hi}\) and \(k_{hj}\) are the \(i\)-th query and \(j\)-th key
(vectors of dimension \(d\)), \(m(h)\) is the head-specific fixed slope (as in
ALiBi), and \(g(i,j)\) is given by \eqtref{eq:scalealibi_2}:

\begin{equation}\label{eq:scalealibi_2}
    g(i,j) = \text{distance(i,j)} \cdot \text{GSD}
\end{equation}

\noindent
where \(distance(i,j)\) is the Euclidean distance between the patches
corresponding to \(i\) and \(j\) and is multiplied with a bias of the GSD of
the source image (which differentiates this approach from vanilla
2D-ALiBi/X-ALiBi). As with ALiBi and 2D-ALiBi/X-ALiBi, the bias is
added during the attention calculation before the softmax step and no attention
is added at the bottom of the network.

\begin{figure}[h]
	\centering
	\includegraphics[width=0.165\textwidth]{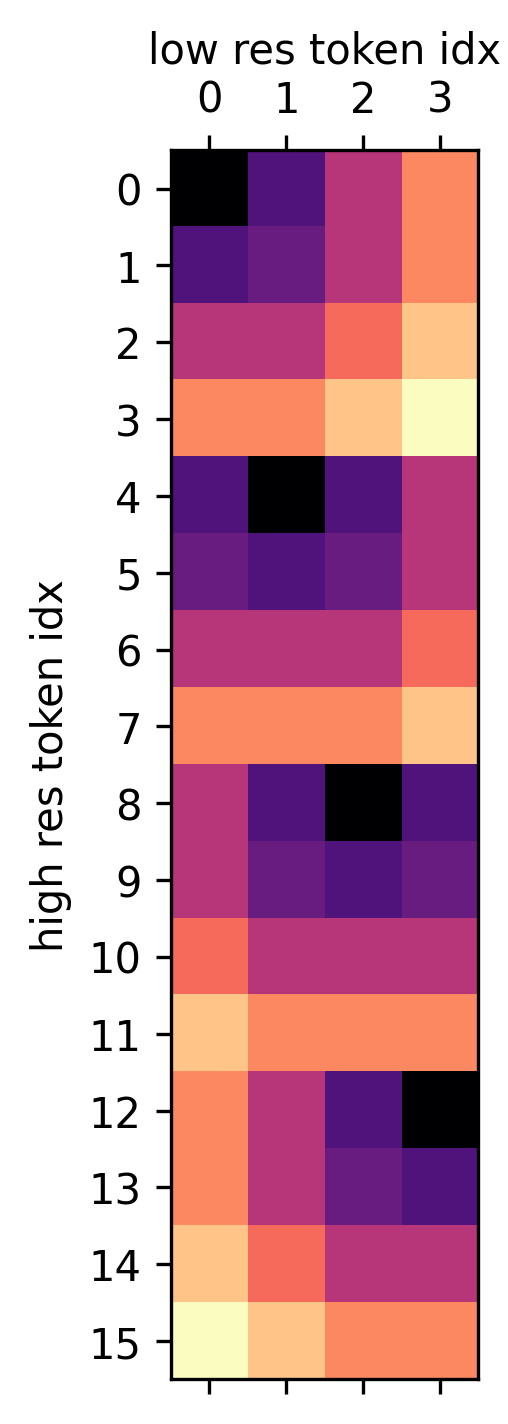}
    \caption{
        An example Scale-ALiBi attention matrix for \(4\) patches of size \(4\) computed
        from a \(4 \times 4\) source image \(s\), with a \(8 \times 8\)
        context image \(c\) containing \(16\) patches. \(s\) and \(c\)
        represent the same physical area on the ground, and thus this matrix
        functions as a distance lookup table comparing these two token
        streams. Note that here the slopes for the different attention heads
        were omitted for clarity.
    }
	\label{fig:cross-attention-matrix}
\end{figure}

This approach allows for the comparison of images at different resolutions,
where a higher GSD (and higher resolution) image may be split into more tokens
than a lower GSD image. For example, a \(256 \times 256\) Sentinel-2 image
sample can be split into \(1024\) patches of size \(8\) whereas a \(512 \times
512\) NAIP image can be split into \(4092\) patches of equivalent size. These
samples represent the same physical area, so a cross-encoder attention can be
encoded as in Figure~\ref{fig:cross-attention-matrix}.

\subsection*{Contrastive learning}

To evaluate the effectiveness of the Scale-ALiBi attention, a setup broadly
similar to CROMA\cite{CROMA} is used, with the addition a high-resolution
encoder \(\Ehires\) and a second cross-encoder to incorporate the
high-resolution token stream into the final images. See Figure~\ref{fig:arch}
for an overview of the model architecture. 

For the contrastive learning step, three separate
optical and SAR ViT encoders are trained: two low-resolution encoders for
the Sentinel-1 synthetic-aperture radar and Sentinel-2 optical observations
(\(\Eradar, \Elores\)) and one high-resolution (2x resolution) encoder for the
aligned NAIP imagery (\(E_\text{hires}\)). Note that the \(\Ehires\) encoder
produces quadruple the number of tokens as the image is at double resolution.
These encoded representations are aligned using an extension of the
standard \mbox{InfoNCE} contrastive loss objective function\cite{infonce} to
add an extra representation, as in \eqtref{eq:infonce_3}:

\begin{equation}\label{eq:infonce_3}
    \mathcal L_\text{Con} = \frac{-1}{|\binom{M}{2}|^2 N}
    \Bigg(\sum^{\binom{M}{2}}_{(m_1,m_2)} \sum^N_i \frac{\exp(z^{i\;\top}_{m_1}
    z^i_{m_2} / \sigma)}{\sum^N_j \exp(z^{j\;\top}_{m_1}
    z^j_{m_2} / \sigma)}\Bigg)
\end{equation}

\noindent
where \(M\) is the set of modalities (in our case low-res.~optical,
high-res.~optical, and SAR), \(\binom{M}{2}\) is the set of 2-combinations of
\(M\), \(N\) is the batch size, \(\sigma\) is the temperature, and \(z_m\) is
the linearized and normalized representation of modality \(m\). When \(\mathcal
L_\text{Con}\) is minimized, the representations of all three modalities from a
single physical location are ``squeezed'' together, and ``pushed'' away from
all other representations of other scenes.

Additionally, the encoded output tokens from the three encoders are
cross-encoded, first with \(\Elores\) and \(\Eradar\) tokens to create a
joint radar-optical encoding, and then this encoding with a second
cross-encoder with the \(\Ehires\) token stream in order to form the final
token stream.

Finally, these tokens are encoded using a masked autoencoder\cite{MAE}
mechanism using a similar setup to CROMA, where the MAE
reconstructs all sensor modalities into a single patch with \(N\) channels,
where \(N\) is the sum of all input channels---effectively fusing the input
sensors\cite{CROMA}. The reconstruction loss \(\mathcal L_\text{Recon}\) is
modified to add a term for the hires token strema, as in~\eqtref{eq:l_recon}:

\begin{equation}\label{eq:l_recon}
    \mathcal L_\text{Recon} = \frac 1 N \sum^N_i \Bigg( \frac{R(t_\text{radar})}{M}
    + \frac{R(t_\text{lores})}{M}+\frac{R(t_\text{hires})}{M} \Bigg)
\end{equation}

\noindent
where \(N\) is the batch size, \(M\) is the number of masked patches, and
\(R(t)\) is sum of the differences between the ground truth \(I_\text{mode}\)
and the mode channels of the
predicted packed representations \(f_\text{dec}(t_\textit{mode})\). As in
CROMA, the encoded tokens are normalized to a mean of \(0\) and a standard
deviation of \(1\). The full function \(R(t)\) is given by~\eqtref{eq:l_recon_R}:

\begin{equation}\label{eq:l_recon_R}
    R(t_\text{mode}) = \sum^M_j I_\text{mode} -\text{Norm}(f_\text{DEC}(t_\text{mode}))
\end{equation}

\noindent
Again, like CROMA, \(f_\text{DEC}(\,\cdot\,)\) is a ViT with a 2D sinusoidal
embedding operating over the masked multimodal patch embeddings. 

The final loss to be optimized is a simple addition of
the contrastive loss and the reconstructive loss, as shown by
\eqtref{eq:final_loss}:

\begin{equation}\label{eq:final_loss}
    \mathcal L = \mathcal L_\text{Con} +
    \mathcal L_\text{Recon}
\end{equation}

\section*{Dataset curation}

In order to train this model, a dataset of paired low-resolution optical,
low-resolution SAR, and high-resolution optical images is required. No existing
public dataset was found that fit the bill, so in addition to the analysis of
the Scale-ALiBi attention, a goal of this research is to curate this dataset.

\begin{figure}
	\centering
    \includegraphics[width=0.4385\textwidth]{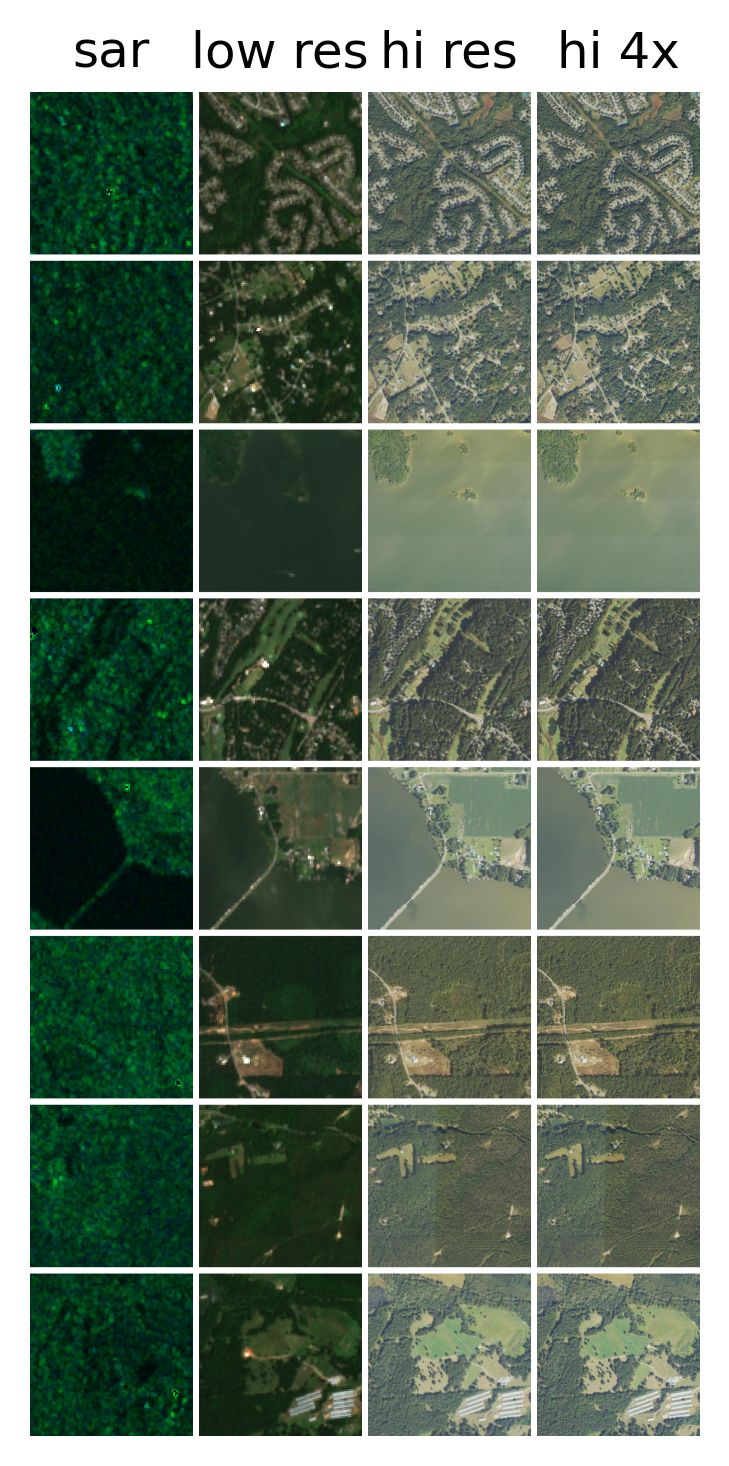}
    \vspace{0.2cm}
    \caption{A selection of samples from the Scale-ALiBi dataset. Note that the rightmost column is double the size of the normal samples.}
	\label{fig:dataset_sample}
\end{figure}

As shown in Figure~\ref{fig:dataset_sample}, in order to generate this dataset Sentinel-1 and Sentinel-2 images are
ingested and segmented into XYZ tiles at a specified level \(Y\) and stored as
PNGs with a \(256 \times 256\) pixel resolution. The true-color image (TCI)
product is segmented from Sentinel-2's L2A collection directly, while the
Sentinel-1 L2A VV and VH captures are scaled to \(\frac {256} {1000}\), with
the VV band assigned to the green channel and VH to the blue channel. An empty
red channel is inserted, and the image is quantized to 8 bits. Then, high
resolution images from NAIP are sourced for the same XYZ tiles at \(Y\) in
order to form a \(256 \times 256\) high-resolution sample. Additionally, the
next tile level down (\(Y+1\)) is collected from NAIP in order to form a \(512
\time 512\) double-resolution image.

Due to the geographic constraints of the NAIP tileset, the Scale-ALiBi dataset
is limited to images covering the continental United States and Puerto Rico. Within
this area, a series of smaller regions are selected for coverage based on
geographic diversity and zoom scale, and these
subsets are released as different dataset sizes. See
Table~\ref{table:dataset_sizes}
for more information about available datasets. Instructions for accessing these is available from the project
website\footnote{\href{https://github.com/pkage/scale-alibi}{https://github.com/pkage/scale-alibi}}.

\begin{table}[h!]
    \caption{Available dataset sizes.}
    \centering
    \begin{tabular}{lllr}
        \toprule
        Name     & Description     & Base \(Y\) & Samples \\
        \midrule
        \texttt{small} & Test/debug set  & 15 &  21,497  \\
        \texttt{full}  & Full size dataset & 15 & 146,502   \\
        \texttt{micro}  & Zoomed-in dataset & 17 & 188,060   \\
        \bottomrule
    \end{tabular}
    \label{table:dataset_sizes}
\end{table}

\section*{Results}

This model is evaluated over the GEO-Bench\cite{geobench} benchmark dataset,
which contains 6 classification and 6 segmentation tasks over both high- and
low-resolution optical and SAR imagery and includes subsets of the datasets for
those tasks. In order to maintain a fair comparison with CROMA while keeping
computational constraints in mind, the CROMA model is trained with identical
data on identical hardware for an equivalent amount of time; and while the preliminary
Scale-ALiBi results fall somewhat short of CROMA's published state-of-the-art results the
authors are optimistic that with a much larger training run, equivalent results
can be achieved. Both of these models were trained with the \(Y=15\) full size dataset (see Table~\ref{table:dataset_sizes}).

For the classification tasks, a neural network with one hidden layer (of size
2048) is used on top of the learned cross-modal representations, as is standard
for representation learning tasks. Additionally, non-parametric methods are
evaluated over the raw represetations, namely \(k\)-means clustering and
\(k\)-nearest neighbors (\(n=20\)). Additionally, a UMAP\cite{UMAP}
dimensionality reduction preprocessing step for the \(k\)-means clustering was
evaluated. Both the high resolution and low resolution optical encoders were
used for the Scale-ALiBi benchmarks, with the benchmark patches being scaled to
\(256 \times 256\) for the low-resolution encoder and \(512 \times 512\) for
the high-resolution encoder. The CROMA benchmark was run identically, except
with the omission of the high-resolution encoder. Overall, Scale-ALiBi
performed similarly or better than CROMA in these benchmarks, with full results
found in Table~\ref{table:benchmarks}.

\begin{table}[h!]
    \caption{GEO-Bench Benchmarks.}
    \centering
    \begin{tabular}{lrrr}
        \toprule
        Name     & SA-high & SA-low   & CROMA \\
        \cmidrule(r){1-4}
        \multicolumn{4}{c}{Neural} \\
        \midrule
        \texttt{m-pv4ger}      & \( 83.27\% \) & \(87.77\%\) & \( \mathbf{88.53\%}\) \\
        \texttt{m-forestnet}   & \( 23.51\%\) & \(27.83\%\) & \(\mathbf {28.27\%}\) \\
        \texttt{m-euronet}     & \( {23.92\% } \) & \( {31.48\% } \) & \( \mathbf{49.04\% } \) \\
        \texttt{m-brick-kiln}  & \( { 70.27\% }  \) & \( {70.07\% } \) & \( \mathbf{75.13\% } \) \\
        \cmidrule(r){1-4}
        \multicolumn{4}{c}{\(k\)-Means} \\
        \midrule
        \texttt{m-pv4ger}      & \( { 50.00\% }  \) & \( \mathbf{52.83\% } \) & \( {50.07\% } \) \\
        \texttt{m-forestnet}   & \( {  8.98\% }  \) & \( { 8.28\% } \) & \( \mathbf{10.84\% } \) \\
        \texttt{m-euronet}     & \( { 11.91\% }  \) & \( \mathbf{12.12\% } \) & \( { 8.34\% } \) \\
        \texttt{m-brick-kiln}  & \( { 50.86\% }  \) & \( \mathbf{51.83\% } \) & \( {49.62\% } \) \\
        \cmidrule(r){1-4}
        \multicolumn{4}{c}{\(k\)-Means + UMAP} \\
        \midrule
        \texttt{m-pv4ger}      & \( { 49.57\% }  \) & \( {48.36\% } \) & \( \mathbf{51.65\% } \) \\
        \texttt{m-forestnet}   & \( \mathbf{  9.18\% }  \) & \( { 8.88\% } \) & \( { 8.18\% } \) \\
        \texttt{m-euronet}     & \( {  9.32\% }  \) & \( \mathbf{11.32\% } \) & \( { 9.36\% } \) \\
        \texttt{m-brick-kiln}  & \( \mathbf{ 52.07\% }  \) & \( {45.40\% } \) & \( {51.66\% } \) \\
        \cmidrule(r){1-4}
        \multicolumn{4}{c}{\(k\)-NN} \\
        \midrule
        \texttt{m-pv4ger}      & \( \mathbf{ 92.39\% } \) & \( {91.89\% } \) & \( {92.29\% } \) \\
        \texttt{m-forestnet}   & \( \mathbf{ 38.26\% } \) & \( {37.26\% } \) & \( {35.44\% } \) \\
        \texttt{m-euronet}     & \( { 58.70\% } \) & \( {64.40\% } \) & \( \mathbf{66.30\% } \) \\
        \texttt{m-brick-kiln}  & \( { 75.37\% } \) & \( {74.97\% } \) & \( \mathbf{76.47\% } \) \\
        \bottomrule
    \end{tabular}
    \label{table:benchmarks}
\end{table}

\section*{Conclusion}

In this paper, we present developmental and preliminary results from the
Scale-ALiBi linear bias attention mechanism for multi-modal and multi-scale
remote sensing foundation models. We provided a reference implementation of the
attention as an extension of CROMA where a high-resolution encoder is added.
This initial model is then benchmarked against an equivalently-trained CROMA
instance, showing a modest improvement. Additionally, a dataset of aligned
low-resolution SAR, low-resolution optical, and high-resolution optical image
sample also of value for remote sensing work and is released alongside the
paper. Future work for this project includes the curation and release of a much
larger dataset, as well as longer training runs for the Scale-ALiBi model.


\vspace{3pt}
\bibliographystyle{ieeetr}
\titleformat{\section}[runin]{\normalsize\bfseries}{\thesection}{0em}{\addperiod}
{\footnotesize \bibliography{references}}

\end{document}